\pdfoutput=1

\documentclass[11pt]{article}

\usepackage[preprint]{acl}

\usepackage{times}
\usepackage{latexsym}
\usepackage{mathbbol}

\usepackage[T1]{fontenc}

\usepackage[utf8]{inputenc}

\usepackage{microtype}

\usepackage{inconsolata}

\usepackage{thmbox}
\newtheorem[M, bodystyle=\noindent, vskip=0.6em]{definition}{Definition}
\usepackage{amsmath}
\usepackage{graphicx}
\usepackage{booktabs}
\usepackage{multirow}
\usepackage{makecell}
\usepackage{adjustbox}
\usepackage{enumitem}
\usepackage{mdframed}
\usepackage[most]{tcolorbox}
\usepackage{float}
\usepackage{bbding}
\usepackage{amssymb} 

\newtcolorbox{prompt}[2][]{
    lower separated=false,
    colback=white,
    boxrule=0.1mm,
colframe=white,
fonttitle=\bfseries,
coltitle=black,
enhanced,
attach boxed title to top left={xshift=0.0cm,
        yshift=-3mm},
boxed title style={size=scriptsize,colback=white, colframe=white},
title=#2,#1}

\usepackage{tikz}

\title{Assessing In-context Learning and Fine-tuning \\ for Topic Classification of German Web Data}

\author{Julian Schelb$^{1}$ \and Roberto Ulloa$^{2}$ \and Andreas Spitz$^{1}$ \\
        $^{1}$Department of Computer Science \\
        $^{2}$Cluster of Excellence ``The Politics of Inequality''  \\
        University of Konstanz \\
        Konstanz, Germany \\
        \texttt{\{julian.schelb, roberto.ulloa, andreas.spitz\}@uni-konstanz.de} \\
}

\begin{document}
\maketitle


\begin{abstract}

Researchers in the political and social sciences often rely on classification models to analyze trends in information consumption by examining browsing histories of millions of webpages. Automated scalable methods are necessary due to the impracticality of manual labeling.
In this paper, we model the detection of topic-related content as a binary classification task and compare the accuracy of fine-tuned pre-trained encoder models against in-context learning strategies. Using only a few hundred annotated data points per topic, we detect content related to three German policies in a database of scraped webpages. We compare multilingual and monolingual models, as well as zero and few-shot approaches, and investigate the impact of negative sampling strategies and the combination of URL \& content-based features.
Our results show that a small sample of annotated data is sufficient to train an effective classifier. Fine-tuning encoder-based models yields better results than in-context learning. Classifiers using both URL \& content-based features perform best, while using URLs alone provides adequate results when content is unavailable.

\end{abstract}


\section{Introduction}

Text classification of webpages is used to understand information consumption by categorizing large collections of individuals' browsing histories (e.g.,~\citealt{stier_post_2022}). By categorizing webpages, researchers can identify patterns of online news consumption~\cite{flaxman_filter_2016} and quantify exposure to populist sentiments~\cite{stier_populist_2022}. Analyzing browsing histories by topic often necessitates "finding the needle in the haystack", as typically just a small fraction of webpage visits correspond to a given domain, such as news sources~\cite{doi:10.1177/19401612211009160}. Therefore, identifying the few relevant pages among numerous unrelated visits makes manual labeling impractical. Machine learning classifiers are often used as an automated and scalable alternative~\cite{stier_populist_2022}. 

Since the introduction of the transformer architecture, fine-tuning pre-trained language models (PLMs) such as BERT~\cite{devlin_bert_nodate} has seen widespread adoption in text classification tasks. Applications include classifying public opinions about policies in digital media~\cite{viehmann_investigating_2023} and identifying protest-related content in newspaper articles~\cite{re_team_2021, sebok_multiclass_2021}. Further applications encompass sentiment analysis on social media posts~\cite{manias_multilingual_2023} and advertising~\cite{jin_combining_2017}. However, fine-tuning classifiers still requires hundreds to thousands of manually labeled documents. Given the multilingual nature of the web and the noisy data resulting from the scraping process, compiling a representative training set remains a complex and time-consuming task. Generative models such as Llama~\cite{touvron_llama_2023} and Mistral~\cite{jiang_mistral_2023} are often inherently multilingual and can generalize to completely unseen tasks without the need for fine-tuning, potentially making them a promising alternative.

In this study, we investigate the use of large language models (LLMs) for the task of binary topic classification across a corpus of scraped webpages. We evaluate our approach by identifying webpages that provide information on three specific German policies discussed during data collection: (1) a policy introduced to combat child poverty, (2) the promotion of renewable energy, and (3) the amendment of cannabis legislation. We compare the classification accuracy between multilingual~\cite{conneauUnsupervisedCrosslingualRepresentation2020} and monolingual~\cite{chan_germans_2020} pre-trained language models by fine-tuning them on manually labeled data. Our analysis extends to generative models~\cite{touvron_llama_2023, chung_scaling_2022}, evaluating few-shot prompting for document classification and assessing the impact of demonstrator sampling strategies.


\section{Related Work}

Political and social sciences researchers increasingly use topic classification to filter large collections of webpages derived from browsing histories~\cite{guess_almost_2021, stier_post_2022}. This task is commonly modeled as binary or multiclass classification, assigning text segments to one or more predefined categories. Until recently, researchers in these applied fields have relied on traditional NLP methods such as naive Bayes classifiers~\cite{stier_post_2022} and logistic regression models~\cite{guess_almost_2021}.

The adaptation of BERT models created new opportunities by improving classification accuracy. For instance, \citet{viehmann_investigating_2023} fine-tuned BERT models to classify opinions on policies in digital media. Similarly, \citet{re_team_2021} explored the use of BERT variants for classifying sentences in newspaper articles to detect protest-related content. \citet{osnabrugge_cross-domain_2023} applied a logistic regression model for classifying the topics of parliamentary speeches. Research on webpage classification also includes the use of URL features~\cite{kan_fast_2005}, extracted content~\cite{jin_combining_2017}, graph representations~\cite{wu_boosting_2015}, and visual features~\cite{xu_new_2015}.

\subsection{Feature-based Learning}
Historically, text classification involved feature engineering by (1) extracting a vector representation of the text, followed by (2) feeding the extracted features into a classifier to determine the final label. Support vector machines~\cite{dorazio_separating_2014} and naive Bayes models~\cite{scharkow_thematic_2013}, often combined with frequency-based tf-idf vectors, were the standard tools. More recently, approaches also rely on techniques such as Word2Vec~\cite{mikolov_efficient_2013} and GloVe~\cite{pennington_glove_2014}, to obtain dense representations of vocabulary items.

\subsection{Contextualized Embeddings}
Recent advancements in text classification have been driven by models like BERT~\cite{devlin_bert_nodate} based on the transformer architecture, which utilize attention mechanisms~\cite{vaswani_attention_nodate} and are trained on extensive unlabeled text datasets through unsupervised pre-training prior to fine-tuning on downstream tasks such as document classification. For instance, mBERT was pre-trained on data from Wikipedias in 104 languages.
XLM-RoBERTa~\cite{conneauUnsupervisedCrosslingualRepresentation2020}, a multilingual extension of RoBERTa~\cite{liu_roberta_2019}, is pre-trained on text from 100 languages. Subsequent fine-tuning of BERT models by replacing the last layer with a classification head for the final prediction has become a common approach~\cite{re_team_2021, gnehm_text_2020, viehmann_investigating_2023, manias_multilingual_2023}.

\begin{table*}[ht]
\small
\centering
\begin{tabular*}{\textwidth}{l@{\extracolsep{\fill}}rrrrrrrrrrrr}
\toprule
\multirow{2}{*}{\textbf{Dataset}} & \multicolumn{2}{c}{\textbf{Children}} & \multicolumn{2}{c}{\textbf{Energy}} & \multicolumn{2}{c}{\textbf{Cannabis}} & \multicolumn{2}{c}{\textbf{All Topics}} \\
\cmidrule(lr){2-3} \cmidrule(lr){4-5} \cmidrule(lr){6-7} \cmidrule(lr){8-9}
 & \textbf{Related} & \textbf{Total} & \textbf{Related} & \textbf{Total} & \textbf{Related} & \textbf{Total} & \textbf{Related} & \textbf{Total} \\
\midrule
\textbf{Training} & 192 & 384 & 204 & 408 & 205 & 410 & 601 & 1,202 \\
\textbf{Unbalanced Test} (Unbl) & 22 & 3,722 & 23 & 4,164 & 23 & 3,448 & 68 & 11,334 \\
\textbf{Balanced Test} (Test) & 22 & 44 & 23 & 46 & 23 & 46 & 68 & 136 \\
\textbf{Extended Test} (Extd) & 45 & 53,253 & 32 & 45,925 & 29 & 44,432 & 106 & 143,610 \\
\midrule
\textbf{Complete Test} (All =  Unbl \& Extd) & 67 & 56,975 & 55 & 50,089 & 52 & 47,880 & 174 & 154,944 \\
\textbf{Complete} (Train, Unbl, \& Extd) & 259 & 57,359 & 259 & 50,497 & 257 & 48,290 & 775 & 156,146 \\
\bottomrule
\end{tabular*}
\caption{\textbf{Number of topic-related and total webpages per topic.} Training and test set contain URLs with high-confidence labels. The unbalanced test set (unbl) includes additional negative examples not included in the training set, while the extended test set (extd) uses low-confidence labels for evaluation under less ideal conditions.}
\vspace{-0.5cm}
\label{tab:dataset-overview}
\end{table*}

\subsection{Models Pre-trained on German Texts}
A considerable amount of research has been dedicated to exploring text classification tasks specifically for the German language~\cite{viehmann_investigating_2023, scharkow_thematic_2013}. Although not all recent studies utilize transformer models for German text classification~\cite{graef_leveraging_2021}, the majority of research underscores the superiority of BERT models in this domain~\cite{gnehm_text_2020}. DBMDZ BERT is comparable in size to BERT-base but is trained on the German segments of the OPUS corpus and Wikipedia. GBERT~\cite{chan_germans_2020} is another German BERT variant that outperforms multilingual models and other German-trained BERT variants~\cite{idrissi-yaghir_domain_2023, niklaus_lextreme_2023, bornheim_fhac_2021}. GBERT includes additional data and implements training enhancements~\cite{chan_germans_2020}, as does the GELECTRA model~\cite{clark_electra_2020}, which is designed for more efficient learning by enabling the model to learn from entire sentences, rather than just the masked tokens. 

\subsection{In-context Learning}
Large generative models like FLAN~\cite{chung_scaling_2022}, Mistral~\cite{jiang_mistral_2023}, and LLaMa~\cite{touvron_llama_2023} are also transformer-based but use stacked decoder blocks instead of the encoder blocks used by BERT. Encoder blocks extract dense vector representations, used as features for classification tasks. Decoder blocks predict the next token to generate output sequences, allowing these models to perform different tasks due to their flexible output schema.

Generative models have demonstrated remarkable generalization across a broad spectrum of NLP tasks by incorporating the instruction directly into the input prompt, often alongside a few labeled examples, thereby eliminating the need for parameter updates. Due to their large training corpora, generative models typically possess some multilingual capabilities. For instance, FLAN is a model family based on the T5 model architecture~\cite{chung_scaling_2022}, able to follow instructions in multiple languages, including English, German, and French. Larger models, like those based on the LLaMA~\cite{touvron_llama_2023} architecture, are further optimized through reinforcement learning from human feedback~\cite{ouyang_training_2022, bai_training_2022}, improving cross-domain generalization and reasoning skills. Aya~\cite{ustun_aya_2024} and Vicuna are further examples. The former is trained on 101 languages including German, while the latter is fine-tuned on user-shared conversations, primarily in English.\footnote{\href{https://sharegpt.com}{https://sharegpt.com}} 

While neural networks have become the state-of-the-art text classification approach, current research lacks a thorough evaluation of LLMs for identifying topic-related content on German webpages. Here, we provide a comprehensive study to fill this gap, including a comparison to traditional feature-based approaches.


\section{Dataset}

For our experiments, we use a corpus of scraped webpages annotated by topic. We describe the data collection and annotation process in Section \ref{subsec:data_coll_ann}. The topic labels correspond to three German policies that were of interest during the period of data collection: (1) basic child support policy (Kindergrundsicherung), introduced to combat child poverty, (2) energy transition policy (Förderung erneuerbarer Energien), designed to promote renewable energy, and the (3) cannabis legalization amendment (Cannabislegalisierung). We refer to these policies as the \emph{children}, \emph{energy}, and \emph{cannabis} policies throughout this paper. 
Our dataset contains substantially more topic-unrelated than relevant webpages. This exemplifies a common challenge in the social, political, and communication sciences: finding relevant content within a vast database of unrelated webpages.

\begin{figure*}
\centering
\includegraphics[width=1.0\textwidth]{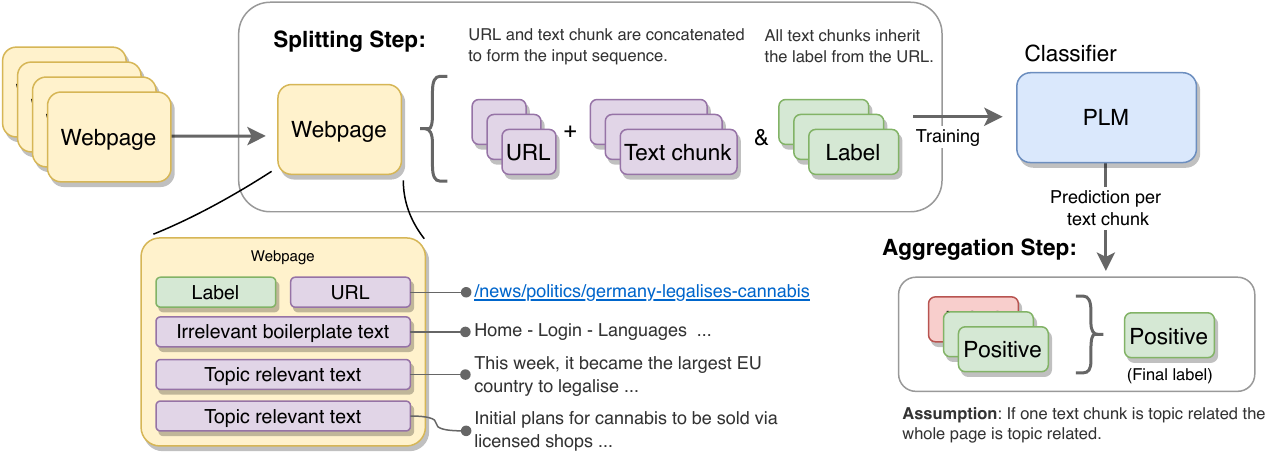}
\caption{ \textbf{Webpage processing and classification pipeline.} The extracted webpage content is divided into chunks, maintaining the original labels. Chunk level predictions are aggregated to obtain the final label per URL.}
\label{fig:processing_pipeline}
\end{figure*}

\subsection{Data Collection and Annotation}
\label{subsec:data_coll_ann}

The browsing traces are obtained as part of a broader project in which $1,228$ participants of a commercial web-tracked panel take part in an online experiment, during which they are instructed to inform themselves about the three policy topics (see Appendix~\ref{appendix:policy_descriptions} and~\ref{appendix:unique_urls_per_category} for details). In total, the participants visit $267$k quasi-unique URLs. Given that only $1,324$ unique URLs ($775$ after filtering) are annotated as policy-related across the three topics, a research assistant augments our training data by manually searching the web for further policy-related webpages. An additional $297$ high-quality positive cases are added for each topic in this way (77, 83, and 137, respectively, for the topics \textit{children}, \textit{energy}, and \textit{cannabis}).

Data from the collected URLs is scraped using the Python package requests\footnote{\href{https://pypi.org/project/requests/}{https://pypi.org/project/requests/}} and the plain text content is extracted from the HTML using the Python package selectolax.\footnote{\href{https://pypi.org/project/selectolax}{https://pypi.org/project/selectolax}}

For each of the three topics, the browsing trace data are manually annotated with binary labels (topic-related or non-relevant) at the URL level. Given the amount of data, we employ a multi-level filtering and refinement approach, moving from hostname categories down to hostnames and finally individual URLs, at each step removing non-relevant URLs. For details on the annotation procedure, see Appendix~\ref{appendix:policy_descriptions}.

After annotation of the successfully scraped webpages ($156$k out of $267$k URLs), our high-confidence data set is comprised of 214 (\textit{children}), 227 (\textit{energy}), 228 (\textit{cannabis}) webpages that are related to the respective topic, and 4,106 (\textit{children}), 4,572 (\textit{energy}), 3,857 (\textit{cannabis}) non-relevant webpages. As a result of the multi-level annotation strategy, we also obtain 143k additional URLs with low-confidence labels that are predominantly negative cases (e.g., searches, YouTube videos, and social media posts), which we use in our evaluation of a real-world application scenario of classifying noisy web data. For further ablative testing on noisy data, we also construct an extended test set with low-confidence labels.

\subsection{Data Preprocessing}
\label{subsec:data_preprocessing}

We describe the processing steps for compiling the datasets for training and evaluation, including sampling train and test examples, as well as segmenting long webpages. We filter out cases where we were unable to retrieve the content, to allow for a 1-to-1 comparison of classification performance based on URLs alone versus using content as an additional feature.

\paragraph{Training and Test Sets.} We partition the dataset for each topic into training and test sets, allocating 90\% of the positive examples to training and 10\% to testing, resulting in three datasets for three binary classification tasks (see Table \ref{tab:dataset-overview}). Only URLs with high-confidence labels are used for the training and test sets (see Section \ref{subsec:data_coll_ann}). The positive cases added during manual augmentation are used exclusively for training.

For our initial experiments, we aim for an even proportion of positive and negative cases in the training and test sets (we discuss suitable sampling strategies in Section~\ref{subsec:sampling}). Further negative examples that are not included are assigned to a second, unbalanced test set (unbl) consisting of predominantly negative examples. This second data set mirrors the original proportion of topic-related and unrelated webpages in our data but still contains only high-confidence URLs. Finally, to assess the performance of the classifiers under real-world conditions, we construct an extended test (extd) set comprised of low-confidence labels. This test set also includes difficult-to-scrape webpages, such as search engines, often resulting in non-useful HTML content due to disabled JavaScript. This dataset is even more unbalanced, containing an overwhelming number of negative cases.

\paragraph{Document Splitting.} 
\label{subsec:document_splitting}
Due to the limited context window of the test LLMs (see Table \ref{tab:model_list}), we divide webpage content into chunks using a recursive text splitter\footnote{\href{https://python.langchain.com/docs/modules/data_connection/document_transformers/}{https://python.langchain.com/docs/}}. 
We utilize a maximum chunk size of 384 tokens for all models, including an overlap of 64 tokens. For each chunk, we assign the label of the parent URL.

\begin{table*}
\small
\centering
\resizebox{\textwidth}{!}{%
\begin{tabular}{lrrrrr}
\toprule
\textbf{Model} & \textbf{Type} & \textbf{Layers} & \textbf{Param.} & \textbf{Languages} & \textbf{Context Size} \\
\midrule 
Multilingual BERT-Base ~\cite{devlin_bert_nodate} & BERT & 12 & 179M & 104 & 512 \\
XLM-RoBERTa-Base~\cite{conneauUnsupervisedCrosslingualRepresentation2020} & RoBERTa & 12 & 279M & 100 & 512 \\
XLM-RoBERTa-Large~\cite{conneauUnsupervisedCrosslingualRepresentation2020} & RoBERTa & 24 & 561M & 100 & 512 \\
German-BERT-Base (\href{https://www.deepset.ai/german-bert}{deepset.ai/german-bert}) & BERT & 12 & 111M & 1 & 512 \\
GELECTRA-Base~\cite{chan_germans_2020} & ELECTRA & 12 & 110M & 1 & 512 \\
GELECTRA-Large~\cite{chan_germans_2020} & ELECTRA & 24 & 336M & 1 & 512 \\
GBERT-Base~\cite{chan_germans_2020} & BERT & 12 & 111M & 1 & 512 \\
GBERT-Large~\cite{chan_germans_2020} & BERT & 24 & 337M & 1 & 512 \\
\midrule 
Aya 101~\cite{ustun_aya_2024} & mT5 & 40 & 13B & 101 & 1024 \\
Vicuna 7b~\cite{vicuna2023} & Llama & 32 & 7B & 1 & 2048 \\
Vicuna 13b~\cite{vicuna2023} & Llama & 40 & 13B & 1 & 2048 \\
FLAN-T5-Base~\cite{chung_scaling_2022} & T5 & 12 & 250M & 60 & 512 \\
FLAN-T5-Large~\cite{chung_scaling_2022} & T5 & 24 & 780M & 60 & 512 \\
FLAN-T5-XXL~\cite{chung_scaling_2022} & T5 & 24 & 11B & 3 & 512 \\
\bottomrule
\end{tabular}
}
\caption{Encoder models used for fine-tuning (top) and generative models used for in-context learning (bottom).}
\label{tab:model_list}
\vspace*{-0.5cm}
\end{table*}


\section{Methods}

We model the detection of topic-related content as a binary classification task for each of the three topics. We compare the F1-scores of fine-tuned encoder models (supervised) and in-context learning strategies (few/zero-shot) against suitable baselines. Figure~\ref{fig:processing_pipeline} shows a schematic overview of the supervised training and classification pipeline. The evaluated LLMs are listed in Table \ref{tab:model_list}. We make the code for our experiments publicly available.\footnote{\href{https://github.com/julianschelb/Topic-Classification}{https:/github.com/julianschelb/Topic-Classification}}

For supervised fine-tuning of monolingual and multilingual models, we experiment with using URL-based features on their own and in combination with content. Due to the small number of webpages related to the three topics, we also experiment with different strategies to sample from the large number of negative examples. For in-context learning classification methods, we evaluate multiple models in zero- and few-shot scenarios, comparing different task demonstrator sampling strategies for the latter.

To aggregate the predicted labels for chunks into document level labels during inference, we assign a positive label to webpages if the label of at least one chunk is predicted to be topic-relevant.

\subsection{Sampling Negative Examples}
\label{subsec:sampling}

To address the imbalance of negative and positive examples in our dataset, we investigate three sampling strategies for negative training examples.

\paragraph{Random.} We select a random subset of webpages classified as negative, aiming for an even number of topic-related and unrelated webpages in our training dataset.

\paragraph{Stratified.} To prevent an overrepresentation of webpages from frequent domains, we group them into strata based on their domain, selecting the 128 most frequent URLs for individual groups and consolidating all remaining ones into a 'others' group.

\paragraph{Cluster-based.} Like~\citealt{sun_text_2023}, we test KNN sampling. We create document vectors using TF-IDF with a dimensionality of 10,000, which we then reduce to 100 dimensions using PCA. Given the unknown total number of clusters, we utilize DBSCAN for clustering and sample webpages from each cluster, including the noise cluster.


\begin{table*}[ht]
\centering
\small
\resizebox{\textwidth}{!}{%
\begin{tabular}{l@{\hspace{0.5em}}l@{\hspace{0.5em}}r@{\hspace{0.5em}}r@{\hspace{0.5em}}r@{\hspace{0.5em}}r@{\hspace{1.2em}}r@{\hspace{0.5em}}r@{\hspace{0.5em}}r@{\hspace{0.5em}}r@{\hspace{1.2em}}r@{\hspace{0.5em}}r@{\hspace{0.5em}}r@{\hspace{0.5em}}r}
\toprule
 & \textbf{Model} & \multicolumn{4}{c}{\textbf{Children}} & \multicolumn{4}{c}{\textbf{Energy}} & \multicolumn{4}{c}{\textbf{Cannabis}} \\
 &  & Test\phantom{*} & Unbl\phantom{*} & Extd\phantom{*} & All\phantom{*} & Test\phantom{*} & Unbl\phantom{*} & Extd\phantom{*} & All\phantom{*} & Test\phantom{*} & Unbl\phantom{*} & Extd\phantom{*} & All\phantom{*} \\
\midrule
\multirow{11}{*}{\rotatebox{90}{\textbf{URL only}}} 
    & Multiling. BERT-Base  & \textbf{0.976}\phantom{*} & 0.205\phantom{*} & 0.023* & 0.032* & 0.958\phantom{*} & 0.072\phantom{*} & 0.007\phantom{*} & 0.013\phantom{*} & \textbf{1.000}\phantom{*} & 0.556* & 0.691* & 0.627* \\
    & XLM-RoBERTa-Base  & 0.900\phantom{*} & 0.141\phantom{*} & \textbf{0.063}* & \textbf{0.076}* & 0.933\phantom{*} & 0.103* & 0.016* & 0.027* & \textbf{1.000}\phantom{*} & 0.541* & 0.533* & 0.536* \\
    & XLM-RoBERTa-Large & \textbf{0.976}\phantom{*} & 0.408* & 0.028* & 0.040* & 0.978\phantom{*} & 0.126* & 0.014* & 0.023* & \textbf{1.000}\phantom{*} & 0.597* & 0.577* & 0.585* \\
    & German-BERT-Base & \textbf{0.976}\phantom{*} & \textbf{0.435*} & 0.030* & 0.042* & \textbf{0.979}\phantom{*} & 0.127* & 0.011\phantom{*} & 0.020\phantom{*} & \textbf{1.000}\phantom{*} & \textbf{0.769}* & 0.422* & 0.522* \\
    & GELECTRA-Large & \textbf{0.976}\phantom{*} & 0.274* & 0.023* & 0.032* & 0.909\phantom{*} & 0.118* & \textbf{0.059}* & \textbf{0.076}* & \textbf{1.000}\phantom{*} & 0.460* & \textbf{0.700*} & 0.575* \\
    & GELECTRA-Base & \textbf{0.976}\phantom{*} & 0.127\phantom{*} & 0.007\phantom{*}  & 0.012\phantom{*} & 0.898\phantom{*} & 0.077\phantom{*} & 0.005* & 0.014\phantom{*} & 0.950\phantom{*} & 0.252\phantom{*} & 0.113\phantom{*} & 0.173\phantom{*} \\
    & GBERT-Large & 0.952\phantom{*} & 0.310* & 0.025* & 0.035* & 0.978\phantom{*} & \textbf{0.173*} & 0.015* & 0.025* & \textbf{1.000}\phantom{*} & 0.755* & 0.667* & \textbf{0.701}* \\
    & GBERT-Base & 0.930\phantom{*} & 0.190\phantom{*} & 0.019\phantom{*} & 0.027\phantom{*} & 0.978\phantom{*} & 0.135* & 0.015* & 0.025* & \textbf{1.000}\phantom{*} & 0.396\phantom{*} & 0.532* & 0.456* \\
    & SVM (Baseline) & 0.950\phantom{*} & 0.174\phantom{*} & 0.017\phantom{*} & 0.024\phantom{*} & 0.898\phantom{*} & 0.072\phantom{*} & 0.012\phantom{*} & 0.019\phantom{*} & 0.947\phantom{*} & 0.321\phantom{*} & 0.185\phantom{*} & 0.223\phantom{*} \\
    & LIB (Baseline) & 0.872\phantom{*} & 0.169\phantom{*} & 0.000\phantom{*} & 0.006\phantom{*} & 0.864\phantom{*} & 0.130\phantom{*} & 0.002\phantom{*} & 0.015\phantom{*} & 0.950\phantom{*} & 0.225\phantom{*} & 0.005\phantom{*} & 0.025\phantom{*} \\
\midrule
& Average (w/o baseline) & 0.958\phantom{*} & 0.261\phantom{*} & 0.027\phantom{*} & 0.037\phantom{*} & 0.951\phantom{*} & 0.116\phantom{*} & 0.018\phantom{*} & 0.028\phantom{*} & 0.994\phantom{*} & 0.541\phantom{*} & 0.529\phantom{*} & 0.522\phantom{*} \\
\midrule
\multirow{10}{*}{\rotatebox{90}{\textbf{URL \& content}}} 
& Multiling. BERT-Base & \textbf{1.000}\phantom{*} & 0.269* & 0.166* & 0.190* & 0.958\phantom{*} & 0.096* & 0.014* & 0.023* & \textbf{0.976}\phantom{*} & 0.556* & 0.304* & 0.375* \\
    & XLM-RoBERTa-Base & \textbf{1.000}\phantom{*} & 0.271* & 0.155* & 0.181* & 0.957\phantom{*} & 0.144* & 0.034* & 0.050* & \textbf{0.976}\phantom{*} & 0.597* & 0.386* & 0.453* \\
    & XLM-RoBERTa-Large & \textbf{1.000}\phantom{*} & 0.323* & 0.287* & 0.298* & 0.957\phantom{*} & 0.168* & 0.030* & 0.045* & \textbf{0.976}\phantom{*} & 0.571* & 0.487* & 0.519* \\
    & German-BERT-Base & \textbf{1.000}\phantom{*} & 0.368* & 0.198* & 0.234* & \textbf{1.000}\phantom{*} & 0.136* & 0.020* & 0.033* & \textbf{0.976}\phantom{*} & 0.440* & \textbf{0.747}* & \textbf{0.578}* \\
    & GELECTRA-Large & \textbf{1.000}\phantom{*} & \textbf{0.500*} & \textbf{0.636*} & \textbf{0.583}* & 0.978\phantom{*} & 0.175* & \textbf{0.136*} & \textbf{0.151}* & \textbf{0.976}\phantom{*} & \textbf{0.625*} & 0.514* & 0.555* \\
    & GELECTRA-Base & \textbf{1.000}\phantom{*} & 0.412* & 0.228* & 0.268* & 0.957\phantom{*} & 0.109* & 0.049* & 0.064* & 0.952\phantom{*} & 0.381* & 0.487* & 0.436* \\
    & GBERT-Large & \textbf{1.000}\phantom{*} & 0.494* & 0.410* & 0.434* & 0.979\phantom{*} & 0.146* & 0.058* & 0.080* & 0.952\phantom{*} & 0.191* & 0.157* & 0.170* \\
    & GBERT-Base & \textbf{1.000}\phantom{*} & 0.333* & 0.249* & 0.272* & 0.957\phantom{*} & \textbf{0.221*} & 0.105* & 0.136* & \textbf{0.976}\phantom{*} & 0.526* & 0.455* & 0.482* \\
    & \text{SVM (Baseline)} & 0.933\phantom{*} & 0.059\phantom{*} & 0.015\phantom{*} & 0.022\phantom{*} & 0.885\phantom{*} & 0.064\phantom{*} & 0.010\phantom{*} & 0.017\phantom{*} & 0.930\phantom{*} & 0.088\phantom{*} & 0.030\phantom{*} & 0.043\phantom{*} \\
\midrule
    & Average (w/o baseline) & 1.000\phantom{*} & 0.371\phantom{*} & 0.291\phantom{*} & 0.308\phantom{*} & 0.968\phantom{*} & 0.149\phantom{*} & 0.056\phantom{*} & 0.073\phantom{*} & 0.970\phantom{*} & 0.486\phantom{*} & 0.442\phantom{*} & 0.446\phantom{*} \\
\midrule
\end{tabular}
}
\caption{F1-score performance of supervised fine-tuning approaches for different feature combinations. Statistical significance is assessed using McNemar's test ($p < 0.05$) with respect to the SVM baseline, denoted by *.}
\label{tab:f1_scores}
\end{table*}

\begin{table*}[ht]
\centering
\small
\resizebox{\textwidth}{!}{%
\begin{tabular}{l@{\hspace{0.2em}}l@{\hspace{3em}}r@{\hspace{0.5em}}r@{\hspace{0.5em}}r@{\hspace{0.5em}}r@{\hspace{1.5em}}r@{\hspace{0.5em}}r@{\hspace{0.5em}}r@{\hspace{0.5em}}r@{\hspace{1.5em}}r@{\hspace{0.5em}}r@{\hspace{0.5em}}r@{\hspace{0.5em}}r}
\toprule
 & \textbf{Sampling Strategy} & \multicolumn{4}{c}{\textbf{Children}} & \multicolumn{4}{c}{\textbf{Energy}} & \multicolumn{4}{c}{\textbf{Cannabis}} \\
 &  & Test & Unbl & Extd & All & Test & Unbl & Extd & All & Test & Unbl & Extd & All \\
\midrule
\multirow{3}{*}{\rotatebox{90}{\textbf{}}} 
 & Random & \textbf{1.000} & \textbf{0.318} & \textbf{0.248} & \textbf{0.268} & \textbf{0.978} & 0.134 & 0.060 & 0.079 & \textbf{0.976} & 0.357 & 0.384 & 0.372 \\
 & Stratified & \textbf{1.000} & 0.300 & 0.156 & 0.185 & \textbf{0.978} & \textbf{0.232} & \textbf{0.112} & \textbf{0.145} & \textbf{0.976} & \textbf{0.548} & \textbf{0.538} & \textbf{0.542} \\
 & Cluster-based & 0.977 & 0.264 & 0.112 & 0.139 & \textbf{0.978} & 0.167 & 0.062 & 0.086 & \textbf{0.976} & \textbf{0.548} & 0.444 & 0.482 \\
\midrule
 & \textbf{Average} & 0.992 & 0.294 & 0.172 & 0.197 & 0.978 & 0.178 & 0.078 & 0.103 & 0.976 & 0.484 & 0.455 & 0.465 \\
\midrule
\end{tabular}%
}
\caption{F1-Score performance of different sampling strategies for GELECTRA-Large}
\label{tab:sampling_strategies_evaluation_best_model}
\vspace*{-0.5cm}
\end{table*}

\subsection{Supervised Classification}

We evaluate several monolingual encoder models that are pre-trained specifically on German texts, as well as multilingual encoder models that include at least a portion of German text in their pre-training data. For fine-tuning, we use the same parameters across all models: a learning rate of $2 \times 10^{-5}$ over a maximum of $3$ epochs. We use a warm-up of $500$ steps at the beginning of training and a weight decay of $0.01$. 

We train one URL-based classifier and one combined URL \& content classifier per topic. Since URLs often contain parts of the article title, categories, or search engine optimization (SEO) keywords, we expect them to be useful for classification~\cite{aljofey_effective_2022, kan_fast_2005}. To avoid overfitting on specific domains only the path and parameter sections of the URL are utilized (see Figure \ref{fig:processing_pipeline}).

\paragraph{Baselines.} For URL-based classification, we use linear interpolation and backoff (LIB) as the baseline~\cite{abramson_2012}. For URL \& content classification, we use support vector machine (SVM) classifiers with TF-IDF vectors for feature extraction, similar to what is frequently employed in the literature~\cite{idrissi-yaghir_domain_2023, kan_fast_2005, dorazio_separating_2014}.

\begin{figure}[ht]
\centering
\includegraphics[width=0.49\textwidth]{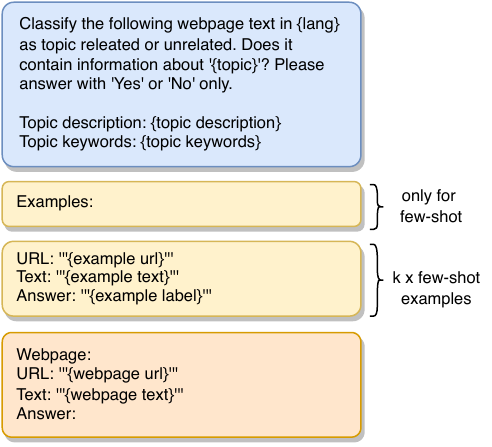}
\caption{\textbf{Prompt template for zero- and few-shot classification.} General task instruction and the incomplete example are consistent across all experiments. For few-shot experiments, $k$ additional demonstrators are included (see Appendix~\ref{appendix:policy_descriptions} for details).}
\label{fig:prompt_template}
\vspace{-0.5cm}
\end{figure}

\subsection{Zero- and Few-Shot Classification}

We evaluate multiple generative models using in-context learning for classification tasks in both zero-shot and few-shot scenarios. We include Aya~\cite{ustun_aya_2024} and two Vicuna variants~\cite{vicuna2023}, as well as three FLAN-T5 variants~\cite{chung_scaling_2022} to assess the performance scaling with model size. Due to the limited context window of FLAN-T5, we evaluate them exclusively in a zero-shot setting. Due to the long inference times, we opted to only evaluate on the balanced test set. Our prompts combine a task description with "Yes" or "No" response instructions to simplify the parsing of the output. Figure \ref{fig:prompt_template} shows the used prompt template. We convert responses to lowercase to map the models' output more easily to a binary label. For answer generation, we set the \texttt{temperature} to 0.3, \texttt{top\_k} to 50, and \texttt{top\_p} to 0.95. While the generative models tend to have longer context windows and would allow for larger webpage chunks, we use the same chunks as the supervised classification for comparison. 

\begin{table*}[ht]
\centering
\small
\begin{tabular}{l@{\hspace{1em}}l@{\hspace{1em}}r@{\hspace{1em}}r@{\hspace{1em}}r@{\hspace{1.5em}}r@{\hspace{1em}}r@{\hspace{1em}}r@{\hspace{1.5em}}r@{\hspace{1em}}r@{\hspace{1em}}r@{\hspace{1.5em}}r@{\hspace{1em}}r@{\hspace{1em}}r}
\toprule
\textbf{} & \textbf{Model} & \multicolumn{3}{c}{\textbf{Children}} & \multicolumn{3}{c}{\textbf{Energy}} & \multicolumn{3}{c}{\textbf{Cannabis}} & \multicolumn{3}{c}{\textbf{All Topics}} \\
\textbf{} &  & Prec & Rec & F1 & Prec & Rec & F1 & Prec & Rec & F1 & Prec & Rec & F1 \\
\midrule
\multirow{6}{*}{\rotatebox{90}{\textbf{Zero-Shot}}} & Aya 101 & \textbf{1.000} & 0.761 & 0.865 & \textbf{1.000} & 0.783 & 0.878 & \textbf{1.000} & 0.950 & 0.974 & \textbf{1.000} & 0.831 & 0.906 \\
 & Vicuna 13b & \textbf{1.000} & 0.714 & 0.833 & \textbf{1.000} & 0.739 & 0.850 & \textbf{1.000} & 0.800 & 0.889 & \textbf{1.000} & 0.751 & 0.857 \\
 & Vicuna 7b & 0.905 & \textbf{0.905} & \textbf{0.905} & 0.950 & 0.826 & 0.884 & \textbf{1.000} & \textbf{1.000} & \textbf{1.000} & 0.952 & \textbf{0.910} & \textbf{0.930} \\
 & FLAN-T5-XXL & \textbf{1.000} & 0.762 & 0.865 & \textbf{1.000} & \textbf{0.870} & \textbf{0.930} & \textbf{1.000} & 0.900 & 0.947 & \textbf{1.000} & 0.844 & 0.914 \\
 & FLAN-T5-Large & 0.944 & 0.810 & 0.872 & 0.938 & 0.652 & 0.769 & \textbf{1.000} & 0.450 & 0.621 & 0.961 & 0.637 & 0.754 \\
 & FLAN-T5-Base & 0.529 & 0.429 & 0.474 & 0.553 & 0.913 & 0.689 & 0.475 & 0.950 & 0.633 & 0.519 & 0.764 & 0.599 \\
\midrule
\multirow{3}{*}{\rotatebox{90}{\shortstack{\textbf{\tiny{Few-Shot}} \\ \textbf{\tiny{Random}}}}}
 & Aya 101 & 0.952 & 0.952 & 0.952 & \textbf{1.000} & 0.870 & 0.930 & 0.905 & 0.950 & 0.927 & 0.952 & 0.924 & 0.936 \\
 & Vicuna 13b & 0.913 & \textbf{1.000} & \textbf{0.955} & \textbf{1.000} & \textbf{0.957} & \textbf{0.978} & \textbf{0.952} & \textbf{1.000} & \textbf{0.976} & \textbf{0.955} & \textbf{0.986} & \textbf{0.970} \\
 & Vicuna 7b & \textbf{1.000} & 0.905 & \textbf{0.955} & 0.512 & \textbf{0.957} & 0.667 & \textbf{0.952} & \textbf{1.000} & \textbf{0.976} & 0.821 & 0.954 & 0.866 \\
\midrule
\multirow{3}{*}{\rotatebox{90}{\shortstack{\textbf{\tiny{Few-Shot}} \\ \textbf{\tiny{Balanced}}}}} & Aya 101 & \textbf{1.000} & 0.762 & 0.865 & \textbf{1.000} & 0.826 & 0.905 & 0.792 & \textbf{0.950} & 0.864 & 0.931 & 0.846 & 0.878 \\
 & Vicuna 13b & \textbf{1.000} & \textbf{1.000} & \textbf{1.000} & \textbf{1.000} & 0.870 & \textbf{0.930} & \textbf{1.000} & \textbf{0.950} & \textbf{0.974} & \textbf{1.000} & \textbf{0.940} & \textbf{0.968} \\
 & Vicuna 7b & \textbf{1.000} & 0.905 & 0.950 & 0.629 & \textbf{0.957} & 0.759 & \textbf{1.000} & \textbf{0.950} & \textbf{0.974} & 0.876 & 0.937 & 0.894 \\
\midrule
\multirow{3}{*}{\rotatebox{90}{\shortstack{\textbf{\tiny{Few-Shot}} \\ \textbf{\tiny{KNN}}}}}
 & Aya 101 & \textbf{0.833} & \textbf{0.952} & \textbf{0.889} & 0.667 & \textbf{0.957} & 0.786 & 0.714 & \textbf{1.000} & 0.833 & 0.738 & \textbf{0.970} & 0.836 \\
 & Vicuna 13b & 0.800 & \textbf{0.952} & 0.870 & \textbf{0.700} & 0.913 & \textbf{0.792} & \textbf{0.952} & \textbf{1.000} & \textbf{0.976} & \textbf{0.817} & 0.955 & \textbf{0.879} \\
 & Vicuna 7b & 0.588 & \textbf{0.952} & 0.727 & 0.524 & \textbf{0.957} & 0.677 & 0.588 & \textbf{1.000} & 0.741 & 0.567 & \textbf{0.970} & 0.715 \\
\midrule
\end{tabular}
\caption{Evaluation of zero-shot learning and few-shot demonstrator sampling strategies on the balanced test set.}
\label{tab:evaluation_results_sampling_strategies}
\end{table*}

\paragraph{Demonstrator Sampling.} Since the selection of task demonstrators included in the few-shot prompt affects prediction quality~\cite{liu_what_2021, peng_revisiting_2024}, we evaluate multiple sampling strategies: (1) random sampling over the training set, (2) random sampling with balanced classes to address class imbalance by ensuring equal representation of each class, and (3) KNN-based sampling, which selects training examples similar to the input~\cite{sun_text_2023}. We calculate the cosine distance based on embeddings extracted using a sentence-transformer~\cite{reimers_sentence-bert_2019}.


\section{Results and Discussion}

\subsection{Supervised Classification Results}

We evaluate all models using URL-only and URL \& content as features and report the F1 scores for the three test datasets (test, unbalanced, and extended) and three topics in Table~\ref{tab:f1_scores}. 

GELECTRA-Large, using URL \& content features, achieves the best average F1 score of 0.430 across all topics on the complete test set (see Table \ref{tab:runtime_comparison}), making it the overall best-performing model. Analyzing the results by topic, GELECTRA-Large achieves the best F1 scores of 0.583 for the \textit{children} topic and 0.151 for the \textit{energy} topic. Meanwhile, German-BERT-Base achieves the best score for the \textit{cannabis} topic with an F1 score of 0.578. 

We discuss the impact of feature selection and negative sampling methods and analyze performance differences between monolingual and multilingual models, as well as base and large models.

\paragraph{URL \& content.} While the URL alone can be an adequate feature for many applications, our findings show that integrating webpage content improves classification performance. Across all topics and models, the average F1 score improved by 40.8\% on the complete test set. 

Classifiers on the \textit{children} topic experienced the most notable improvement, with F1 scores increasing by 4.4\% on the test set, 42.1\% on the unbalanced set, an substantial 977.3\% on the extended set, and 731.1\% on the complete set, indicating that content helps the classifier to generalize. The \textit{energy} topic also showed enhanced performance with the inclusion of content features. Interestingly, the \textit{cannabis} topic exhibited a decrease in average performance. This decrease may be attributed to ground truth labels being annotated at the URL level rather than the content level. Webpages on this topic might utilize URLs with highly expressive keywords, enabling the URL-only classifier to perform very effectively. Alternatively, as our manual error analysis suggests (see \ref{subsec:manual_error_analysis}), webpages discussing this topic but lacking topic-relevant keywords in the URLs might have been missed during the annotation process. 

In summary, classifiers trained on URL \& content perform better, especially on the challenging extended test set.

\paragraph{Performance Comparison: Test Sets.} All models perform well on the balanced test set with both URL \& content-based features, but their performance significantly deteriorates on the unbalanced and extended test sets. The average performance across all topics decreases by 65.7\% from the balanced to the unbalanced set and by 73.1\% to the extended set. Although recall remains high, the drop in precision indicates an increase in false positives, confirming the greater difficulty of these datasets due to lower quality scraped content and less reliable labels. The results show that the classifiers struggle with noise in the extracted webpage content introduced by the scraping process.

\begin{figure}[ht]
\centering
\includegraphics[width=0.49\textwidth]{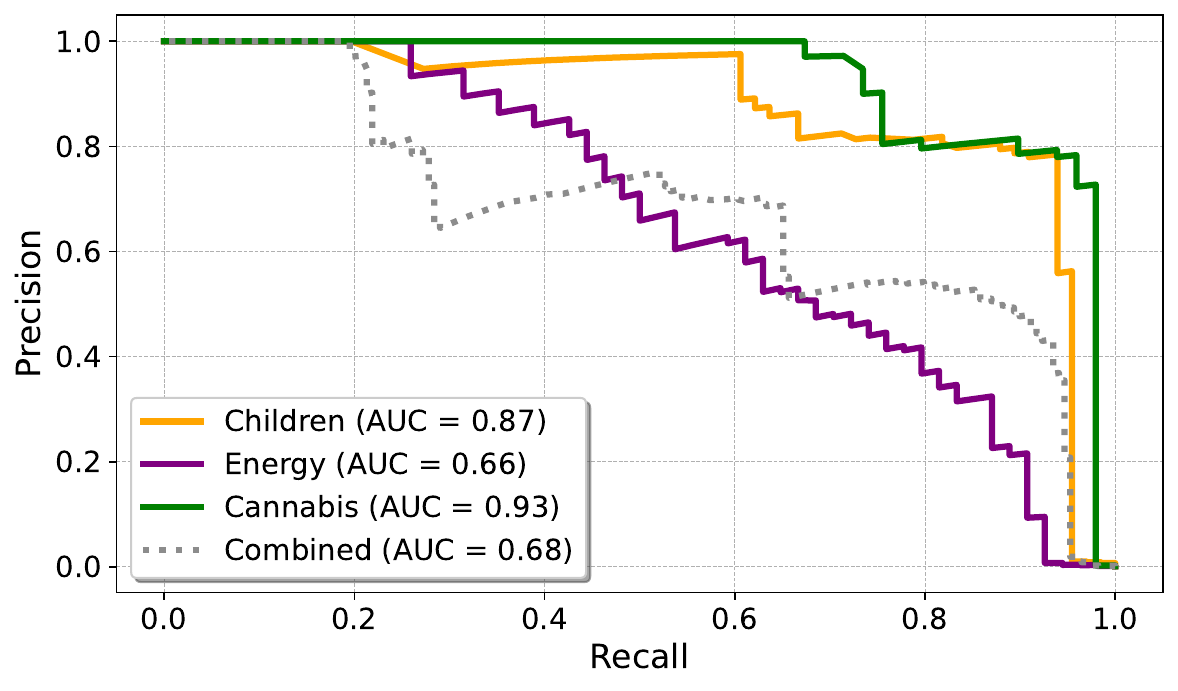}
\caption{\textbf{Precision-recall curves for GELECTRA-Large} across topics on the Complete test set. Cannabis shows the highest precision-recall performance and Energy the lowest (recall that the number of webpages varies between the topics).}
\label{fig:pr-curves}
\end{figure}

\paragraph{Performance Comparison: Topics.} 
\textit{Cannabis}-related webpages are generally the easiest to detect, while \textit{energy}-related webpages are the most challenging. This observation aligns with our intuition, as \textit{cannabis} represents a more specific topic. In contrast, the \textit{energy} topic is considerably broader, overlapping with a range of areas that are unrelated to the topic of renewable energy, such as climate change. The precision-recall curves based on all available data, as depicted in Figure \ref{fig:pr-curves}, further support this observation.

\paragraph{Monolingual vs. Multilingual Models.} Monolingual models achieve a mean F1 score 25.9\% higher than multilingual models on the complete test set across all topics when using URL \& content features. Comparing the best monolingual model, GELECTRA-Large, with the best multilingual model, xlm-roberta-large, GELECTRA-Large achieves an F1 score that is 22.4\% higher on the unbalanced dataset, 60.0\% higher on the extended dataset, and 49.5\% higher on the complete test set.

\paragraph{Negative Sampling.} In Table \ref{tab:sampling_strategies_evaluation_best_model}, we report the results comparing three negative sampling strategies. We find that random sampling and stratified sampling perform comparably, with stratified sampling yielding slightly better performance overall.

\paragraph{Model Size and Runtime Analysis.} Larger models generally outperform their base variants, with modest gains. On the unbalanced dataset, the average F1 score increases by 9.4\% (from 0.32 to 0.35), while on the extended dataset, scores see a more substantial boost of 25\% (from 0.24 to 0.30). These improvements highlight the benefits of larger models in handling more complex and varied data. However, this increased performance comes at a significant cost in processing time. As shown in Table~\ref{tab:runtime_comparison}, large variants achieve better F1 scores but process only \textasciitilde19 webpage chunks per second, compared to \textasciitilde63 chunks for the base variants. This 28\% gain in F1 score comes with a 200\% increase in processing time. The SVM baseline is the fastest at \textasciitilde1000 chunks per second but has the lowest F1 score. Measurements were conducted using an Nvidia Tesla P100 GPU and an Intel Xeon Gold 6132 CPU @ 3.700GHz.

\begin{table}[ht]
\centering
\small
\resizebox{\columnwidth}{!}{%
\begin{tabular}{l@{\hspace{1em}}r@{\hspace{1em}}r@{\hspace{1em}}r}
\toprule
\textbf{Model} & \textbf{URL} & \textbf{URL\&C} & \textbf{Chunks/sec} \\
\midrule
Multiling. BERT-Base & 0.224 & 0.196 & 59 \\
XLM-RoBERTa-Base & 0.213 & 0.228 & 63 \\
XLM-RoBERTa-Large & 0.216 & 0.287 & 20 \\
German BERT-Base & 0.195 & 0.282 & 67 \\
GELECTRA-Large & 0.228 & \textbf{0.430} & 19 \\
GELECTRA-Base & 0.066 & 0.256 & 63 \\
GBERT-Large & \textbf{0.254} & 0.228 & 19 \\
GBERT-Base & 0.169 & 0.297 & 63 \\
SVM (Baseline) & 0.022 & 0.027 & \textbf{1000} \\
\bottomrule
\end{tabular}
}
\caption{Average F1 scores on the complete test set over the three topics and inference throughput (chunks/sec) averaged over 5 runs on the unbalanced test set.}
\label{tab:runtime_comparison}
\end{table}

\subsection{Zero- and Few-shot Results}

Our results demonstrate that zero-shot and few-shot methods deliver good performance (see Table \ref{tab:evaluation_results_sampling_strategies}). The best zero-shot model, determined by averaging the F1 scores across the three topics, is Vicuna 7b, which achieves an average F1 score of 0.930. The overall best model is Vicuna 13b with few-shot and random sampling of task demonstrators, which achieves an average F1 score of 0.970. For sampling task demonstrators, random and random balanced sampling strategies work better than KNN-based sampling. However, few-shot classification remains consistently inferior to fine-tuning, which is therefore the preferred approach for achieving optimal results if labeled data is available.

\subsection{Manual Error Analysis}
\label{subsec:manual_error_analysis}

We perform a manual error analysis on the predictions of the best performing classifier, GELECTRA-Large with random negative sampling, by randomly sampling 50 misclassified webpage chunks from both the unbalanced and extended test sets per topic, yielding 300 chunks in total. The errors are categorized by type in Table~\ref{table:error_type_overview} (for a more detailed breakdown, see Appendix~\ref{appendix:error_analysis}).

In 42 instances, the classifier's prediction is correct and the ground truth is incorrect (GT error). This is not surprising since the extended test set consists primarily of webpages with low-confidence labels and the manual labeling is URL-based, while the classifier analyzes individual chunks within the scraped content.
In 85 instances, webpage chunks contained very general information pertaining to the topic but were not truly relevant (topic related). Examples include pharmacies selling cannabis, online solar panel shops, and energy price comparison portals. Conversely, in 50 instances, the classifier identified webpages from ministries or institutions discussing other laws as topic-relevant (law related). Both cases highlight the inherent difficulty in distinguishing topical information from specific legal content.
Furthermore, we find that the classifier is sensitive to words like "legal," "Umwelt" (environment), and "Verkehr" (transportation), resulting in 56 misclassified cases (unrelated). Additionally, in 52 cases, the classifier misclassified boilerplate chunks, such as navigation elements or cookie banners, likely because all chunks inherit the webpage's URL-based label (boilerplate). This caused some chunks to be labeled as topic-relevant without containing relevant information, introducing noise to the training dataset.
Finally, in 15 cases, web scraping or preprocessing failed to produce meaningful content, which confused the classifier (content error). Errors include warnings about disabled JavaScript, login-protected content, or encoding issues.

\begin{table}[t]
\small
\centering
\resizebox{\columnwidth}{!}{%
\begin{tabular}{lr p{3.5cm}}
\toprule
\textbf{Error Type} & \textbf{Count} & \textbf{Example URL} \\
\midrule
GT error & 42 & \href{http://sanitygroup.com/}{http://sanitygroup.com/} \\
Topic related & 85 & \href{http://luckyhemp.de}{http://luckyhemp.de} \\
Law related & 50 & \href{https://buergergeld.org}{https://buergergeld.org} \\
Unrelated & 56 & \href{http://gutefrage.net/frage/chef-zahlt-bar-auf-die-hand-legal}{http://gutefrage.net/} \\
Boilerplate & 52 & - \\
Content error & 15 & - \\
\bottomrule
\end{tabular}%
}
\caption{Error analysis of 300 misclassified chunks}
\label{table:error_type_overview}
\vspace*{-0.5cm}
\end{table}


\section{Conclusion}

We compare the performance of fine-tuned encoder models against in-context learning strategies for the classification of topic-related content. Using only a few hundred positively annotated data points per topic, we detect content related to three German policies in a database of scraped webpages. The best supervised classifier, GELECTRA-Large, using URL \& content features, achieves an average F1 score of 0.430 over all topics,  performance varies by topic. It performs well on the \textit{children} and \textit{cannabis} topics but performs suboptimal in terms of precision for the \textit{energy} topic.

All fine-tuned models achieve strong performance on the high-quality balanced test set, regardless of using URL or content-based features. However, performance declines substantially on lower-quality and unbalanced data, with high recall but lower precision due to more pages being falsely labeled as topic-related. While recall remains high across all topics and test sets, precision drops considerably, leading to a substantial number of false positives, which indicates that the model is overly sensitive to keywords that are topic-related but also occur in other contexts. 
Webpage content proved to be a strong signal for classification over URL-based baselines, and classifiers that combined URL \& content-based features perform best. In cases where content-based analysis is infeasible, URL-based classifiers can provide an adequate baseline performance, although the precision-recall tradeoff in settings with real-world data requires a careful approach.
However, a manual error analysis revealed that the classifiers struggle to distinguish between weak and strong relations to the topic, with URL-based labels leading to incorrect associations of boilerplate texts with the topic. An investigation of more elaborate chunk pooling and combination strategies in future work is needed. Additionally, incorporating loosely topic-related negative examples into the training data would likely improve classifier precision by enabling better differentiation between relevant and non-relevant instances. For instance, online shops that advertise cannabis or solar panels are relevant to the topic in general but not in the sense of political policy discussion.

Our evaluation shows high accuracy for zero- and few-shot prompting without fine-tuning, indicating their potential in data-constrained situations. Few-shot learning can be viable when runtime is less critical, but labeled data is expensive. However, fine-tuning encoder-based models generally yields better results and should be given preference over in-context learning for annotating large datasets. 

\paragraph{Future Work.} It is likely that classifier precision can be enhanced by filtering out topic-unrelated chunks and training a content-only classifier to remove unrelated content. To address the limited number of positive examples, data augmentation appears like a fruitful addition to the pipeline. For in-context learning, advanced prompting methods such as prompt chaining and chain-of-thought prompting are likely to enhance LLM reasoning.

\section*{Limitations}

\paragraph{URL-based Labeling.} Since we generated training data based on URL-level labeling of websites as a proxy for content-based labeling for reasons of feasibility, it is likely that our data (and therefore our findings) are biased. While the manual error analysis indicated that just 14\% of errors are ground-truth errors, this amount is non-negligible. In settings where resources are available for proper content-based labeling, it is likely that this error can be reduced.

\paragraph{Website Chunking.} Since we assign URL-level labels to webpage chunks, it is likely that chunks in the training data are labeled incorrectly. As described in Section \ref{subsec:data_preprocessing}, we split webpage content into chunks due to the 512-token input limit for our classifiers, with each chunk inheriting the URL's label. Thus, if a webpage is labeled as topic-relevant, all chunks receive a positive label, even if some contain irrelevant text, such as navigation elements or cookie banners. As a result of this, the model sometimes associates boilerplate text with the positive class. The pragmatic solution here is to go with the times and use models with larger input sizes to avoid chunking altogether.

\paragraph{Scraping-induced Noise.} Another source of noise stems from the web scraping process. For example, our web scraper did not support JavaScript, causing many webpages to display warnings or malfunction. In these cases, the URL label remains positive, indicating topic-related content, but the scraper failed to retrieve that content, further introducing noise in the training data. Similar issues occur with login protected webpages, dynamic content, cookie banners, YouTube videos, and PDFs.

\section*{Ethics Statement}

The browsing traces from which we scraped the web data were provided by Bilendi GmbH, which hosts a web tracking panel. The company adheres to EU GDPR regulations, and participants were fully informed about the data collection process, including the option to temporarily disable tracking for privacy reasons. A letter of information was provided, and consent was requested from all participants upon first contact and then thereafter at each additional contact point. Ethics approval has been received by the University of Konstanz IRB under the number IRB23KN02-003/w.

\section*{AI Policy Statement}

In conducting our research and preparing this paper, we utilized AI assistants, specifically ChatGPT and GitHub Copilot. ChatGPT was employed for paraphrasing and refining the authors' original content to enhance clarity and readability, without suggesting new content. GitHub Copilot assisted in coding tasks by providing code suggestions and completions.

\section*{Acknowledgements}
We owe many thanks to Katharina Jäger and Corinna Nitsch for their help with data annotation, and Anton Pogrebnjak for support in data scraping. We also extend our gratitude to Juhi Kulshrestha and Celina Kacperski for helpful discussions. This research was funded by the Deutsche Forschungsgemeinschaft (DFG – German Research Foundation) under Germany‘s Excellence Strategy – EXC-2035/1 – 390681379. The funders had no role in study design, data collection and analysis, decision to publish, or preparation of the manuscript.

\bibliography{references}

\appendix
\setcounter{table}{0}

\section{Data collection}
\label{appendix:policy_descriptions}

The URLs forming the basis for the corpus of this study were obtained as part of a broader project in which individuals of a commercial web-tracked panel were invited to participate in an online experiment. Participants (N=1228) were randomly assigned to one of 3 groups: a control group, and two intervention groups (both instructed to search about the policy topics, but only one with a financial incentive), with weekly instructions to inform themselves about the three policy topics during a 20-30h window. The visited URLs were recorded (N=~761K), and the content was scraped.

\paragraph{}{Children.} The "Kindergrundsicherung" (basic child support) policy aims to combat child poverty by providing a fixed amount, income-dependent supplement, and educational benefits.\footnote{\href{https://www.bmfsfj.de/bmfsfj/service/gesetze/gesetz-zur-einfuehrung-einer-kindergrundsicherung-und-zur-aenderung-weiterer-bestimmungen-bundeskindergrundsicherungsgesetz-bkg--230650}{https://www.bmfsfj.de/bmfsfj/service/gesetze/gesetz-zur-einfuehrung-einer-kindergrundsicherung-und-zur-aenderung-weiterer-bestimmungen-bundeskindergrundsicherungsgesetz-bkg--230650}}

\paragraph{Energy.} The EEG 2023 (Erneuerbare-Energien-Gesetz, Renewable Energy Sources Act) aims to increase the share of renewable energies in gross electricity consumption to at least 80\% by 2030.\footnote{\href{https://www.bundesregierung.de/breg-de/schwerpunkte/klimaschutz/novelle-eeg-gesetz-2023-2023972}{https://www.bundesregierung.de/breg-de/schwerpunkte/klimaschutz/novelle-eeg-gesetz-2023-2023972}}

\paragraph{Cannabis.} The CanG 2023 (Cannabisgesetz, Cannabis Control Act) will legalize the private cultivation of cannabis by adults for personal use and collective non-commercial cultivation.\footnote{\href{https://www.bundesgesundheitsministerium.de/themen/cannabis/faq-cannabisgesetz}{https://www.bundesgesundheitsministerium.de/themen/ cannabis/faq-cannabisgesetz}}

\section{URL Annotation Process}
\label{appendix:policy_descriptions_b}

During the 20h-30h windows of the experiment, participants visited $\sim 761K$ URLs comprising $\sim 267K$ quasi-unique URLs (i.e., the sum of the total unique URLs per topic). To obtain training examples, the URL annotation protocol followed a multi-level strategy:

\begin{enumerate}
    \item \textbf{Hostname category:} Hostnames (\(N = 17,207\)) were classified according to three categorizations: (1) base categories provided by the commercial panel (\(N=48\)), and (2) the simplified categories (\(N=46\)) and (3) IAB categories (\(N=405\)) gathered via the Webshrinker service. Three researchers (two postdocs and one research assistant) indicated if the base and simplified categories were irrelevant to the topic, i.e., were unlikely to contain policy-related information; two annotators (one postdoc and one research assistant) did so for the IAB categories. Only URLs from unanimously irrelevant categories were discarded.
    \item \textbf{Hostname:} We extracted the unique hostnames corresponding to the remaining URLs (homepages were excluded). One research assistant indicated that the hostname was irrelevant (i.e., unlikely to contain information relevant to the topic). If so, the hostname was discarded. As an exception, the next level directly included URLs corresponding to a curated list of news hostnames (\(N\approx700\), Stier et al., 2020) because they are likely to include topic-related information (so checking those domains manually is unnecessary).
    \item \textbf{URL:} URLs were sorted into categories (see Table 2). URLs that fall into the “Other” category were not annotated (14.7\%) because most would require visiting the URL. One of the authors checked the hostnames and judged them to be not very likely to contain relevant information. One annotator indicated if the remaining URLs were related to the policy topic.
\end{enumerate}

For the experiments in the study, three annotated URL categories were excluded: (1) web searches because the post-hoc scraping would alter the results the participants encounter, (2) social media because the content is not accessible (via scraping), and (3) YouTube because the API was used instead of web-scraping (and the content does not strictly correspond to webpages).

In total, 4983 URLs for \textit{children}, 5782 for \textit{energy}, and 4834 for cannabis manually annotated URLs were used in this study; only 139, 180, and 76, respectively, were relevant to each topic.

\section{Distribution of unique URLs}
\label{appendix:unique_urls_per_category}

The distribution of annotated URLs according to their category and topic is presented in Table \ref{tab:URLCategorization}. During the multistep annotation process, some categories, such as social media and web searches, are discarded before manual analysis due to their unlikely relevance to the topic (see column "Used"). Categories with high-confidence labels (used = yes) include URLs with SEO-optimized titles, news without SEO-optimized titles, Wikipedia, and keyworded domains, while web searches, social media, YouTube shorts and videos, and other miscellaneous URLs have only low-confidence labels (used = no). The latter categories form the basis of our extended test set. The URL counts in Table \ref{tab:URLCategorization} indicate the total number of URLs annotated. The number of webpages in our dataset used in our experiments is lower because cases where content cannot be retrieved using our web scraper are excluded.

\newpage

\begin{table*}[ht]
\centering
\small
\resizebox{\textwidth}{!}{%
\begin{tabular}{l@{\hspace{0.5em}}r@{\hspace{1.2em}}r@{\hspace{1.2em}}r@{\hspace{1.2em}}p{6.3cm}@{\hspace{0.5em}}r}
\toprule
\textbf{URL Category} & \textbf{Children} & \textbf{Energy} & \textbf{Cannabis} & \textbf{Details} & \textbf{Used} \\
\midrule
Web searches & 6723 & 6374 & 7869 & Identified by query search parameters such as the \texttt{q} in \texttt{google.com/search?q=value} & No \\
\midrule
URLs with SEO-optimized title & 3713 & 4476 & 3947 & Identified by hyphenated separation of long strings, such as \textit{example.com/germany-legalises-cannabis}  & Yes \\
\midrule
News without SEO-optimized title & 498 & 559 & 624 & Identified using a manually curated list of news hostnames, such as \textit{example.com} & Yes \\
\midrule
Social Media & 469 & 482 & 529 & Due to GDPR, the provider excludes URLs visited by fewer than 3 people. However, under our request, they included unique visits to lists of media and politicians by HBI and BTW17 & No \\
\midrule
Wikipedia & 208 & 301 & 271 & Wikipedia titles do not follow SEO standards & Yes \\
\midrule
YouTube shorts and videos & 1656 & 1433 & 1875 & YouTube API was used to obtain metadata (e.g., title and description) for the classification & No \\
\midrule
Keyworded Domains & 33 & 182 & 106 & URLs corresponding to domains that contain common keywords identified in the web searches or the SEO titles, such as \textit{example-cannabis-info.com} & Yes \\
\midrule
Other & 1822 & 2750 & 2711 & URLs that does not match any above categories. & No \\
\bottomrule
\end{tabular}
}
\caption{Distribution of unique annotated URLs by category and topic. In addition to the number of unique URLs in each category, we include methodological details about the categorization.}
\label{tab:URLCategorization}
\end{table*}

\section{Manually-augmented data}
\label{appendix:search_augmented_data}

Given the scarcity of topic-relevant URLs among the annotated cases, a research assistant was instructed to complement our training dataset using the Google search engine. Three query terms were based on how the policy topics were referred to in the online survey experiment: \textit{"kindergrundsicherung"}, \textit{"gesetze zur förderung erneuerbarer energien"}, and \textit{“cannabis legalisierung"}. The process was twofold:

\begin{enumerate}
    \item First, the assistant downloaded approximately 15 non-news results related to the topic among the top 30, limiting the search until July 31st, 2023.
    \item Second, they performed nine monthly-restricted news searches between November 1st, 2022, and July 31st, 2023, downloading those relevant to the topic among the top 10 results (top 20 for cannabis).
\end{enumerate}

In total, 77, 83, and 137 webpages were added for each topic, respectively.

\section{Manual Error Analysis}
\label{appendix:error_analysis}

In our manual error analysis of GELECTRA-Large with random negative sampling, we examine 300 misclassified webpage chunks. Identifying these errors helps us refine labeling, enhance preprocessing, and adjust the model to better distinguish relevant from irrelevant content. See Table~\ref{tab:error-analysis} for a detailed breakdown.

This analysis highlights areas for improvement in our model. For instance, in 52 cases, boilerplate text (e.g., navigation elements, cookie banners) is predicted as topic-relevant by the classifier, likely due to URL-based ground truth labels. The 512-token input limit necessitates chunking the webpage content. For URLs with positive labels, all chunks, sometimes including boilerplate, inherit the URL's positive label. This causes the model to associate boilerplate text with the positive class during training. Using models with larger input sizes could mitigate this issue.

Noise from the web scraping process is another concern, as indicated by the 15 examples in our sample. Our web scraper does not support JavaScript, leading to errors when retrieving content from some webpages. This highlights the importance of URL-only classifiers as a fallback.

\begin{table*}[ht]
\centering
\small
\resizebox{\textwidth}{!}{%
\begin{tabular}{l@{\hspace{0.5em}}p{5.5cm}@{\hspace{0.5em}}r@{\hspace{0.5em}}p{5.5cm}}
\toprule
\textbf{Error Type} & \textbf{Error Descriptions} & \textbf{Count} & \textbf{Example URL} \\
\midrule
Ground truth error & The classifier's prediction is correct and the ground truth is incorrect. This is often due to the Extended test set consisting primarily of webpage chunks with low-confidence labels and the manual labeling being URL-based while the classifier analyzes chunks within the scraped content. & 42 & \href{http://sanitygroup.com/}{sanitygroup.com}, \newline \href{http://tecson.de/heizoelpreise.html}{tecson.de/heizoelpreise.html}, \newline \href{http://barth-wuppertal.de/warum-eine-neue-gasheizung-noch-sinn-macht}{barth-wuppertal.de/warum-eine-neue-gasheizung-noch-sinn-macht},  \newline \href{http://kinder-grund-sicherung.de/impressum}{kinder-grund-sicherung.de/impressum}, \newline \href{https://www.cdu.de/artikel/ganzheitliche-loesungen-statt-buerokratie}{cdu.de/artikel/ganzheitliche-loesungen-statt-buerokratie} \\
\midrule
Topic related & Webpage chunks contain general information pertaining to the topic but are not truly relevant. Examples include pharmacies selling cannabis products, online shops selling solar panels, and web portals comparing energy prices. & 85 & 
\href{http://luckyhemp.de}{luckyhemp.de}, \newline \href{http://leafly.de/}{leafly.de}, \newline \href{http://solaridee.de/}{solaridee.de}, \newline \href{https://www.hwk-stuttgart.de/e-mobilitaet}{hwk-stuttgart.de/e-mobilitaet}, \newline \href{https://www.umweltbundesamt.de}{umweltbundesamt.de} , \newline \href{https://hartz4antrag.de/}{hartz4antrag.de/} \\
\midrule
Law related & The classifier identifies webpage chunks from ministries or institutions discussing other laws as policy-relevant. This highlights the difficulty in distinguishing topical information from specific legal content. & 50 & \href{http://landkreisleipzig.de/pressemeldungen.html?pm_id=5477}{landkreisleipzig.de}, \newline \href{http://hartziv.org/}{hartziv.org}, \newline \href{http://leipzig.de/umwelt-und-verkehr}{leipzig.de/umwelt-und-verkehr}, \newline 
\href{http://www.fuehrungszeugnis.bund.de/ffwr}{fuehrungszeugnis.bund.de/ffw}, \newline \href{http://loerrach-landkreis.de/}{loerrach-landkreis.de/} \\
\midrule
Unrelated & The classifier is sensitive to words like "legal," "Umwelt" (environment), and "Verkehr" (transportation), leading to misclassification of irrelevant webpage chunks. & 56 & \href{http://www.lernstudio-barbarossa.de/regensburg}{lernstudio-barbarossa.de/regensburg}, \newline \href{https://www.biker-boarder.de/cannondale/2824204s.html}{biker-boarder.de/cannondale/2824204s.html},
\newline \href{http://kachelmannwetter.com/de/wetteranalyse/hessen}{kachelmannwetter.com/de/wetteranalyse/},
\newline \href{http://swr.de/}{swr.de/} \\
\midrule
Boilerplate & Misclassification of boilerplate chunks, such as navigation elements or cookie banners, due to all chunks inheriting the webpage's URL-based label. This introduces noise into the training dataset. & 52 & - \\
\midrule
Content error & Web scraping or preprocessing failures produce unusable text, confusing the classifier. Errors include warnings about JavaScript, login-protected content, or encoding issues. & 15 & - \\
\bottomrule
\end{tabular}
}
\caption{Categorization of 300 misclassified webpage chunks; sampled from unbalanced and extended test sets}
\label{tab:error-analysis}
\end{table*}

\end{document}